%% file: ijcai21.tex
\pdfoutput=1
\documentclass{article}
\usepackage{ijcai21}

\usepackage{times}
\usepackage{soul}
\usepackage{url}
\usepackage[hidelinks]{hyperref}
\usepackage[utf8]{inputenc}
\usepackage[small]{caption}
\usepackage{graphicx}
\usepackage{amsmath}
\usepackage{amsthm}
\usepackage{booktabs}
\usepackage{algorithm}
\usepackage{algorithmic}
\usepackage{subcaption}
\usepackage{latexsym}

\title{Partition Function Estimation: A Quantitative Study}

\author{
Durgesh Agrawal$^1$
\and
Yash Pote$^2$\And
Kuldeep S. Meel$^{2}$
\affiliations
$^1$Indian Institute of Technology Kanpur, Kanpur, India\\
$^2$National University of Singapore, Singapore
}

\usepackage{todonotes}
\usepackage{multirow}
\usepackage{changepage}
\usepackage{mathtools}
\usepackage{caption}
\usepackage{graphicx}
\usepackage{amsfonts}
\usepackage{tikz}

\newcommand{\tap}{{TAP}}
\newcommand{\specialcell}[2][l]{%
	\begin{tabular}[#1]{@{}l@{}}#2\end{tabular}}

\newtheorem{prop}{Proposition}
\usepackage{hyperref}
 \hypersetup{
	colorlinks=true,
	linkcolor=blue,
	filecolor=blue,
	citecolor = black,      
	urlcolor=cyan,
}

\begin{document}

\maketitle

\input{./sections/abstract.tex}

\input{./sections/intro.tex}

\input{./sections/prelims.tex}

\input{./sections/algorithms.tex}

\input{./sections/tables.tex}

\input{./sections/methodology.tex}
\input{./sections/images.tex}

\input{./sections/evaluation.tex}

\input{./sections/conclusion.tex}
\input{./sections/acks.tex}

{\small
\bibliographystyle{named}
\bibliography{ijcai21}
}
\end{document}

%% file: sections/abstract.tex
\begin{abstract}
    Probabilistic graphical models have emerged as a powerful modeling tool for several real-world scenarios where one needs to reason under uncertainty. A graphical model's partition function is a central quantity of interest, and its computation is key to several probabilistic reasoning tasks. Given the \#P-hardness of computing the partition function, several techniques have been proposed over the years with varying guarantees on the quality of estimates and their runtime behavior. This paper seeks to present a survey of 18 techniques and a rigorous empirical study of their behavior across an extensive set of benchmarks. Our empirical study draws up a surprising observation: exact techniques are as efficient as the approximate ones, and therefore, we conclude with an optimistic view of opportunities for the design of approximate techniques with enhanced scalability. Motivated by the observation of an order of magnitude difference between the Virtual Best Solver and the best performing tool, we envision an exciting line of research focused on the development of portfolio solvers.
\end{abstract}

%% file: sections/intro.tex
\section{Introduction}

\noindent Probabilistic graphical models are ubiquitously employed to capture probabilistic distributions over complex structures and therefore find applications in a wide variety of domains~\cite{D09,KF09,M12}. For instance, image segmentation and recognition can be modeled into a statistical inference problem~\cite{FF08}. In computational protein design, by modeling force fields of a protein and another target molecule as Markov Random Fields and computing partition function for the molecules in bound and unbound states, their affinity can be estimated~\cite{VSBS16}.

Given a probabilistic graphical model where the nodes correspond to variables of interest, one of the fundamental problems is computing the normalization constant or the partition function. Calculation of the partition function is computationally intractable owing to the need for summation over possibly exponentially many terms. Formally, the seminal work of Roth (\citeyear{R96}) established that the problem of computation of partition function is \#P-hard. Since the partition function plays a crucial role in probabilistic reasoning, the development of algorithmic techniques for calculating the partition function has witnessed a sustained interest from practitioners over the years~\cite{MSS09,ML12,G13,MPWX13,WM19}.

The algorithms can be broadly classified into three categories based on the quality of computed estimates:
\begin{enumerate}
	\item \textit{Exact}~\cite{P86,LS88,P88,HSC89,JLO90,SS90,D95,D96,D99,D01J,CDJ04,D04,MD05,D11,LM17}
	\item \textit{Approximate}~\cite{JS93,D95,GD05,GLD12,EGSS13a,EGSS13b,KSE18,LMID19,KDZ19,SRSM19}
	\item \textit{Guarantee-less}~\cite{P82,YFW00,M01,DKM02,WH03,QM04,E09,LI11,KCT20}
\end{enumerate}
While the exact techniques return an accurate result, the approximate methods typically provide $(\varepsilon, \delta)$ guarantees such that the returned estimate is within $\varepsilon$ factor of the true value with confidence of at least $1-\delta$. Finally, the guarantee-less methods return estimates without any accuracy or confidence guarantees.
Another classification of the algorithmic techniques can be achieved based on the usage of underlying core technical ideas:

\begin{enumerate}
	\item \textit{Message passing}~\cite{P82,YFW00,M01,WJW01,DKM02,WH03,QM04,E09}
	\item \textit{Variable elimination}~\cite{D96,D99,LI11,PCGFRSSV19,LMID19}
	\item \textit{Model counting}~\cite{D01,CDJ04,CD07,D11,EGSS13a,EGSS13b,VSBS16,LM17,GAE18,SBZKE18,SRSM19,DM19,WGE19,LM19,SBWE19,WCGE20,ABM20,AM20,SGM20,ZMY20,DFM20}
	\item \textit{Sampling}~\cite{H88,SP89,DFMR00,W02,GD05,EGSS11,GD11,MPWX13,LPIF15,LDI17,D19,SKMG17,CE20,PWK20}.
\end{enumerate}

Given the plethora of techniques, their relative performance may not always be apparent to a practitioner. It may contribute to the non-usage of the state-of-the-art method, thereby limiting the potential offered by probabilistic graphical models. In particular, we draw inspiration from a related sub-field of automated reasoning: SAT solving, where a detailed evaluation of SAT solvers offered by SAT competition informs the practitioners of the state-of-the-art~\cite{HJS19}. An essential step in this direction was taken by the organization of the six UAI Inference challenges from 2008 to 2016~\footnote{\url{www.hlt.utdallas.edu/~vgogate/uai16-evaluation}}. While these challenges have highlighted the strengths and weaknesses of the different techniques, a large selection of algorithmic techniques has not been evaluated owing to a lack of submissions of the corresponding tools to these competitions. 

This survey paper presents a rigorous empirical study spanning 18 techniques proposed by the broader UAI community over an extensive set of benchmarks. To the best of our knowledge, this is the most comprehensive empirical study to understand the behavior of different techniques for computation of partition function/normalization constant. Given that computation of the partition function is a functional problem, we design a new metric to enable runtime comparison of different techniques for problems where the ground truth is unknown. To judge long-running or non-terminating algorithms fairly, we use a two-step timeout that allows them to provide sub-optimal answers. Our proposed metric, {\tap} score, captures both the time taken by a method and its computation accuracy relative to other techniques on a unified numeric scale.

Our empirical study throws up several surprising observations: the weighted counting-based technique, Ace~\cite{CD05} solves the largest number of problems closely followed by loopy and fractional belief propagation. While Ace falls in the category of exact methods, several of the approximate and guarantee-less techniques, surprisingly, perform poorly compared to the exact techniques.  Given the \#P-hardness of computing the partition function, the relatively weak performance of approximate techniques should be viewed as an opportunity for future research. Furthermore, we observe that for every problem, at least one method was able to solve in less than 20 seconds with a 32-factor accuracy. Such an observation in the context of SAT solving led to an exciting series of works on the design of portfolio solvers~\cite{HHS07,XHHL08} and we envision development of such solvers in the context of partition function.  

The remainder of this paper is organized as follows. In Section~\ref{sec:notation}, we introduce the preliminaries while Section~\ref{sec:algorithms} surveys different techniques for partition function estimation. In Section~\ref{sec:methodology}, we describe the objectives of our experimental evaluations, the setup, and the benchmarks used. We present our experimental findings in Section~\ref{sec:results}. The paper is concluded in Section~\ref{sec:conclusion}.

%% file: sections/prelims.tex
\section{Preliminaries}\label{sec:notation}

A graphical model consists of variables and factors. We represent sets in \textbf{bold}, and their elements in regular typeface.

Let $\textbf{X}=\{x_1, x_2, ..., x_n\}$ be the set of discrete random variables. Let us consider a family $\mathbf{S}\subseteq 2^{\mathbf{X}}$. For each $\mathbf{x}\in\mathbf{S}$, we use $f_{\mathbf{x}}$ to denote a factor, which is a function defined over $\mathbf{x}$. $f_{\mathbf{x}}$ returns a non-negative real value for each assignment $\sigma(\mathbf{x})$ where each variable in $\mathbf{x}$ is assigned a value from its domain. In other words, if $\mathbf{D}_{\mathbf{x}}$ denotes the cross product of domains of all variables in $\mathbf{x}$, then $f_{\mathbf{x}} : \mathbf{D}_{\mathbf{x}} \rightarrow \mathbb{R}^{+}$

The probability distribution is often represented as a bipartite graph, called a factor graph $G=(\mathbf{X}\cup \mathbf{S}, \mathbf{E})$, where $(x_i, \mathbf{x})\in \mathbf{E}$ iff $x_i \in \mathbf{x}$.

The probability distribution encoded by the factor graph is $
\mathbb{P}\left(\sigma(\mathbf{X})\right) = \dfrac{1}{Z}\prod\limits_{\mathbf{x}\in\mathbf{S}} f_{\mathbf{x}}\left(\sigma(\mathbf{x})\right)$, where the normalization constant, denoted by $Z$ and also called partition function, is defined as
\begin{align*}
Z:=\sum\limits_{\sigma(\mathbf{X})} \prod\limits_{\mathbf{x}\in\mathbf{S}} f_{\mathbf{x}}\left(\sigma(\mathbf{x})\right)
\end{align*}

We focus on techniques for the computation of $Z$.

%% file: sections/algorithms.tex
\section{Overview of Algorithms}\label{sec:algorithms}

We provide an overview of the central ideas behind the algorithms we have included in this study. The algorithms can be broadly classified into four categories based on their fundamental approach:

\subsection{Message Passing-based Techniques}

Message Passing algorithms involve sending messages between objects that could be variable nodes, factor nodes, or clusters of variables, depending on the algorithm. 
Eventually, some or all the objects inspect the incoming messages to compute a \textit{belief} about what their state should be. These beliefs are used to calculate the value of $Z$.

\noindent{\bfseries Loopy Belief Propagation} was first proposed by Pearl (\citeyear{P82}) for exact inference on tree-structured graphical models~\cite{KFL01}. The sum-product variant of LBP is used for computing the partition function. For general models, the algorithm's convergence is not guaranteed, and the beliefs computed upon convergence may differ from the true marginals. The point of convergence corresponds to a local minimum of Bethe free energy.

\paragraph{Conditioned Belief Propagation}
Conditioned Belief Propagation is a modification of LBP. Initially, Conditioned BP chooses a variable $x$ and a state $X$ and performs Back Belief Propagation (back-propagation applied to Loopy BP) with $x$ \textit{clamped} to $X$ (i.e., conditioned on $x=X$), and also with the negation of this condition. The process is done recursively up to a fixed number of levels. The resulting approximate marginals are combined using estimates of the partition sum~\cite{E09}.

\noindent{\bfseries Fractional Belief Propagation} modifies LBP by associating each factor with a weight. If each factor $f$ has weight $w_f$, then the algorithms minimize the $\alpha$-divergence~\cite{AIS01} with $ \alpha=1/w_f $ for that factor~\cite{WH03}. Setting all the weights to 1 reduces FBP to Loopy Belief Propagation. 

\noindent{\bfseries Generalized Belief Propagation} modifies LBP so that messages are passed from a group of nodes to another group of nodes~\cite{YFW00}. Intuitively, the messages transferred by this approach are more informative, thus improving inference. The convergence points of GBP are equivalent to the minima of the Kikuchi free energy. 

\noindent{\bfseries Edge Deletion Belief Propagation} is an anytime algorithm that starts with a tree-structured approximation corresponding to loopy BP, and incrementally improves it by recovering deleted edges~\cite{CD06}.

\noindent{\bfseries HAK Algorithm}
Whenever Generalised BP~\cite{YFW00} reaches a fixed point, it is known that the fixed point corresponds to the extrema of the Kikuchi free energy. However, generalized BP does not always converge. The HAK algorithm solves this typically non-convex constrained minimization problem through a sequence of convex constrained minimizations of upper bounds on the Kikuchi free energy~\cite{HAK03}.

\noindent{\bfseries Join Tree} partitions the graph into clusters of variables such that the interactions among clusters possess a tree structure, i.e., a cluster will only be directly influenced by its neighbors in the tree. Message passing is performed on this tree to compute the partition function. $Z$ can be exactly computed if the local (cluster-level) problems can be solved in the given time and memory limits. The running time is exponential in the size of the largest cluster~\cite{LS88,JLO90}.

\noindent{\bfseries Tree Expectation Propagation} represents factors with tree approximations using the expectation propagation framework, as opposed to LBP that represents each factor with a product of single node messages. The algorithm is a generalization of LBP since if the tree distribution approximation of factors has no edges, the results are identical to LBP~\cite{QM04}.

\subsection{Variable Elimination-based Techniques}
Variable Elimination algorithms involve eliminating an object (such as a variable or a cluster of variables) to yield a new problem that does not involve the eliminated object~\cite{ZP94}. The smaller problem is solved by repeating the elimination process or other methods such as message passing.      

\noindent{\bfseries Bucket Elimination} partitions the factors into buckets, such that each bucket is associated with a single variable. Given a variable ordering, the bucket associated with a variable $x$ does not contain factors that are a function of variables higher than $x$ in the ordering. Subsequently, the buckets are processed from last to first. When the bucket of variable $x$ is processed, an elimination procedure is performed over its functions, yielding a new function $f$ that does not mention $x$, and $f$ is placed in a lower bucket. The algorithm performs exact inference, and the time and space complexity are exponential in the problem's induced width~\cite{D96}.

\noindent{\bfseries Weighted Mini Bucket Elimination} is a generalization of Mini-Bucket Elimination, which is a modification of Bucket Elimination and performs approximate inference. It partitions the factors in a bucket into several mini-buckets such that at most $iBound$ variables are allowed in a mini-bucket. The accuracy and complexity increase as $iBound$ increases.

Weighted Mini Bucket Elimination associates a weight with each mini-bucket and achieves a tighter upper bound on the partition function based on Holder's inequality~\cite{LI11}.

\subsection{Model Counting-based Techniques}
Partition function computation can be reduced to one of the weighted model counting~\cite{D02}. A factor graph is first encoded into a CNF formula $\varphi$, with an associated weight function $W$ assigning weights to literals such that the weight of an assignment is the product of the weight of its literals. Given $\varphi$ and $W$, computing the partition function reduces to computing the sum of weights of satisfying assignments of $\varphi$, also known as weighted model counting~\cite{CD08}.

\noindent{\bfseries Weighted Integral by Sums and Hashing}(WISH) reduces the problem into a small number of optimization queries subject to parity constraints used as hash functions. It computes a constant-factor approximation of partition function with high probability~\cite{EGSS13a,EGSS13b}.

\noindent{\bfseries d-DNNF based tools} reduce weighted CNFs to a deterministic Decomposable Negation Normal Form (d-DNNF), which supports weighted model counting in time linear in the size of the compiled form~\cite{DM02}.

\textit{Ace}~\cite{CD07} extracts an Arithmetic Circuit from the compiled d-DNNF, which is used to compute the partition function.
\textit{miniC2D}~\cite{OD15} is a Sentential Decision Diagram (SDD) compiler, where SDDs are less succinct and more tractable subsets of d-DNNFs~\cite{D11}.

\noindent{\bfseries GANAK} uses 2-universal hashing-based probabilistic component caching along with the dynamic decomposition-based search method of sharpSAT~\cite{T06} to provide probabilistic exact counting guarantees~\cite{SRSM19}.

\noindent{\bfseries WeightCount} converts weighted CNFs to unweighted~\cite{CFMV15}, and ApproxMC~\cite{SM19} is used as the model counter.

\subsection{Sampling-based Techniques}
Sampling-based methods choose a limited number of configurations from the sample space of all possible assignments to the variables. The partition function is estimated based on the behavior of the model on these assignments.

\noindent{\bfseries SampleSearch} augments importance sampling with a systematic constraint-satisfaction search, guaranteeing that all the generated samples have non-zero weight. When a sample is supposed to be rejected, the algorithm continues instead with a systematic search until a non-zero weight sample is generated~\cite{GD11}.

\noindent{\bfseries Dynamic Importance Sampling}
interleaves importance sampling with the best first search, which is used to refine the proposal distribution of successive samples. Since the samples are drawn from a sequence of different proposals, a weighted average estimator is used that upweights higher-quality samples~\cite{LDI17}.

\noindent{\bfseries FocusedFlatSAT} is an MCMC technique based on the flat histogram method. FocusedFlatSAT proposes two modifications to the flat histogram method: \textit{energy saturation} that allows the Markov chain to visit fewer energy levels, and \textit{focused-random walk} that reduces the number of null moves in the Markov chain~\cite{EGSS11}.

%% file: sections/tables.tex
\begin{table*}[t]
	\centering
	\setlength\tabcolsep{0.7mm}
	
	\small
	\begin{tabular}{lccccccccc}
		\toprule
		\multicolumn{1}{c}{\multirow{3}{*}{\textbf{Method Name}}}  & \multicolumn{9}{c}{\textbf{Problem Classes}}                                                                                                                                                                                                                                                                                                                                                                                                                                                                                                                                                                                                                                                  \\ 
		\multicolumn{1}{c}{}                                      
		& \begin{tabular}[c]{@{}c@{}}Relation-\\al (354)\end{tabular}
		& \begin{tabular}[c]{@{}c@{}}Prome-\\ das (65)\end{tabular}
		& \begin{tabular}[c]{@{}c@{}}BN\\ (60)\end{tabular}
		& \begin{tabular}[c]{@{}c@{}}Ising\\ (52)\end{tabular}
		& \begin{tabular}[c]{@{}c@{}}Segment\\ (50)\end{tabular}

		& \begin{tabular}[c]{@{}c@{}}ObjDetect\\ (35)\end{tabular}
		& \begin{tabular}[c]{@{}c@{}}Protein\\ (29)\end{tabular}

		& \begin{tabular}[c]{@{}c@{}}Misc\\(27)\end{tabular}
		& \begin{tabular}[c]{@{}c@{}}Total\\ (672)\end{tabular}
		\\ \midrule
		
\specialcell{Ace } & 354 & 65 & 60 & 51 & 50 & 0 & 16 & 15 & 611\\

\specialcell{Fractional Belief Propagation (FBP) } & 293 & 65 & 58 & 41 & 48 & 32 & 29 & 9 & 575\\

\specialcell{Loopy Belief Propagation (BP) } & 292 & 65 & 58 & 41 & 46 & 32 & 29 & 10 & 573\\

\specialcell{Generalized Belief Propagation (GBP) } & 281 & 65 & 36 & 47 & 40 & 34 & 29 & 9 & 541\\

\specialcell{Edge Deletion Belief Propagation (EDBP)} & 245 & 42 & 56 & 50 & 49 & 35 & 28 & 23 & 528\\

\specialcell{GANAK } & 353 & 58 & 53 & 4 & 0 & 0 & 7 & 14 & 489\\

\specialcell{Double Loop Generalised BP (HAK) } & 199 & 65 & 58 & 43 & 43 & 35 & 29 & 14 & 486\\

\specialcell{Tree Expectation Propagation (TREEEP) } & 101 & 65 & 58 & 50 & 48 & 35 & 29 & 15 & 401\\

\specialcell{SampleSearch} & 89 & 56 & 33 & 52 & 37 & 35 & 29 & 25 & 356\\

\specialcell{Bucket Elimination (BE) } & 98 & 32 & 15 & 52 & 50 & 35 & 29 & 22 & 333\\

\specialcell{Conditioned Belief Propagation (CBP) } & 109 & 32 & 21 & 41 & 50 & 35 & 29 & 8 & 325\\

\specialcell{Join Tree (JT) } & 98 & 32 & 15 & 52 & 50 & 19 & 26 & 21 & 313\\

\specialcell{Dynamic Importance Sampling (DIS)} & 24 & 65 & 25 & 52 & 50 & 35 & 29 & 27 & 307\\

\specialcell{Weighted Mini Bucket Elimination (WMB) } & 68 & 13 & 17 & 50 & 50 & 20 & 28 & 12 & 258\\

\specialcell{miniC2D } & 187 & 1 & 30 & 31 & 0 & 0 & 0 & 1 & 250\\

\specialcell{WeightCount} & 93 & 0 & 27 & 0 & 0 & 0 & 0 & 0 & 120\\

	\specialcell{WISH} & 0 & 0 & 0 & 9 & 0 & 0 & 0 & 0 & 9\\
	
	\specialcell{FocusedFlatSAT} & 6 & 0 & 0 & 0 & 0 & 0 & 0 & 0 & 6\\

		\bottomrule
	\end{tabular}
	
	\caption{$\#$Problems solved with 32-factor accuracy}  \label{tab:solved}
	
\end{table*}

\begin{table}[ht]
	\setlength{\tabcolsep}{4pt}
	\centering
	\small
	\begin{tabular}[t]{lccccc}
		\toprule
		{Problem}  & {\#bench-} & {$|\mathbf{X}|$} & {$|\mathbf{S}|$} & {Avg. var.} & {Avg. factor}\\
		{class}  & {marks} &  & & {cardinality} & {scope size}\\
		\midrule
		{Relational}     & 354  &  13457  &  14012  &  2.0  &  2.64                       \\
				{Promedas}       & 65  &  639  &  646  &  2.0  &  2.15                       \\
		{BN}             & 60  &  613  &  658  &  2.46  &  2.8                          \\
		
		{Ising}         & 52  &  93  &  270  &  2.0  &  1.67                      \\
		
		{Segment}        & 50  &  229  &  851  &  2.0  &  1.73                       \\

						{ObjDetect}      & 35  &  60  &  200  &  17.14  &  1.7                       \\	
		{Protein}        & 29  &  43  &  115  &  15.81  &  1.58                       \\

		{Misc}         & 27  &  276  &  483  &  2.35  &  2.03              \\
		\bottomrule
	\end{tabular}
	
	\caption{Variation in graphical model parameters over benchmarks for which $\hat{Z}$ is available. Entries in the last 4 columns are medians over the entire class}  \label{tab:problems}
	
\end{table}

%% file: sections/methodology.tex
\section{Experimental Methodology}\label{sec:methodology}
We designed our experiments to rank the algorithms that compute partition function according to their performance on a benchmark suite. A similar exercise is carried out every year where SAT solvers are compared and ranked on the basis of speed and number of problems solved. However, the task of ranking partition function solvers is complicated by the following three factors:
\begin{enumerate}
	\item For a majority of the problems in the benchmark suite, the ground truth is unknown.
	\item Unlike SAT, which is a decision problem, partition function problem is functional, i.e., its output is a real value.
	\item Some solvers gradually converge to increasingly accurate answers but do not terminate within the given time.
\end{enumerate} 

\subsection{Benchmarking}\label{sec:benchmarking}
As the ground truth computation for large instances is intractable, we used the \textit{conjectured} value of $Z$, denoted by $\hat{Z}$, as the baseline which was computed as follows:

\begin{enumerate}	
	\item If either Ace or the Join Tree algorithm could compute $Z$ for a benchmark, it was taken as true $Z$. For these benchmarks, $\hat{Z}=Z$.
	\item For all such benchmarks where $\hat{Z}=Z$, if an algorithm either (a) returns an accurate answer or (b) no answer at all, we called the algorithm \textit{reliable}.
	\item For the benchmarks where an accurate answer was not known, if one or more \textit{reliable} algorithms gave an answer, their median was used as $\hat{Z}$.

\end{enumerate}
By this approach, we could construct a \textit{reliable} dataset of 672 problems.

\subsection{Timeout}
Since many algorithms do not terminate on various benchmarks, we set a timeout of 10000 seconds for each algorithm. In many cases, even though an algorithm does not return an answer before timeout, it can give a reasonably close estimate of its final output based on the computations performed before timeout. To extract the outputs based on incomplete execution, we divided the timeout into two parts:
\begin{enumerate}

	\item \textit{Soft Timeout:} Once the soft timeout is reached, the algorithm is allowed to finish incomplete iteration, compile metadata, perform cleanups, and give an output based on incomplete execution. We set this time to 9500 seconds.
	
	\item \textit{Hard Timeout:} On reaching the hard timeout, the algorithm is terminated, and is said to have timed-out without solving the problem. We set this time to 500 seconds after the soft timeout.
	
\end{enumerate}

\subsection{Comparing Functional Algorithms}

The algorithms vary widely in terms of the guarantees offered and the resources required. We designed a scoring function to evaluate them on a single scale for a fair comparison amongst all of them. The metric is an extension of the PAR-2 scoring system employed in the SAT competition.

\noindent{\bfseries The {\tap} score} or the Time Accuracy Penalty score system gives a performance score for a particular solver on one benchmark.
We define the {\tap} score as follows:
\begin{align*}
\text{\tap}(t,\mathcal{R}) =
\begin{cases}
2\mathcal{T}
\hfill\hfill
\mathrm{hard\;timeout/error/memout}\\
t+ \mathcal{T}\times\mathcal{R}/{32}
\hfill\hfill
\mathrm{\mathcal{R}<32}\\
2\mathcal{T}-(\mathcal{T}-t)\times\exp(32-\mathcal{R})
\hfill\;\;\;\;\;\;\hfill
\mathrm{\mathcal{R}\ge 32}
\end{cases}
\end{align*}
\noindent
where $\mathcal{T}=10000$ seconds is the hard timeout,\\
$t<\mathcal{T}$ is the time taken to solve the problem, and\\
$\mathcal{R}=\mathrm{max}\left(Z_{ret}/\hat{Z},\hat{Z}/Z_{ret}\right)$ is the relative error in the returned value of partition function $Z_{ret}$ with respect to $\hat{Z}$.

\begin{prop}
	The {\tap} score is continuous over the domain of $t$ and $\mathcal{R}$, where $t<10000$ seconds and $\mathcal{R}\ge 1$.
\end{prop}

The score averaged over a set of benchmarks is called the mean {\tap} score and is a measure of the overall performance of an algorithm on the set. It considers the number of problems solved, the speed and the accuracy to give a single score to the solver. For two solvers, the solver with a lower mean {\tap} score is the better performer. Figure \ref{fig:varR} shows the variation in {\tap} score with $\mathcal{R}$ for a constant $t$.

%% file: sections/images.tex
\begin{figure}[t]
	\centering

	\includegraphics[clip,width=0.95\columnwidth]{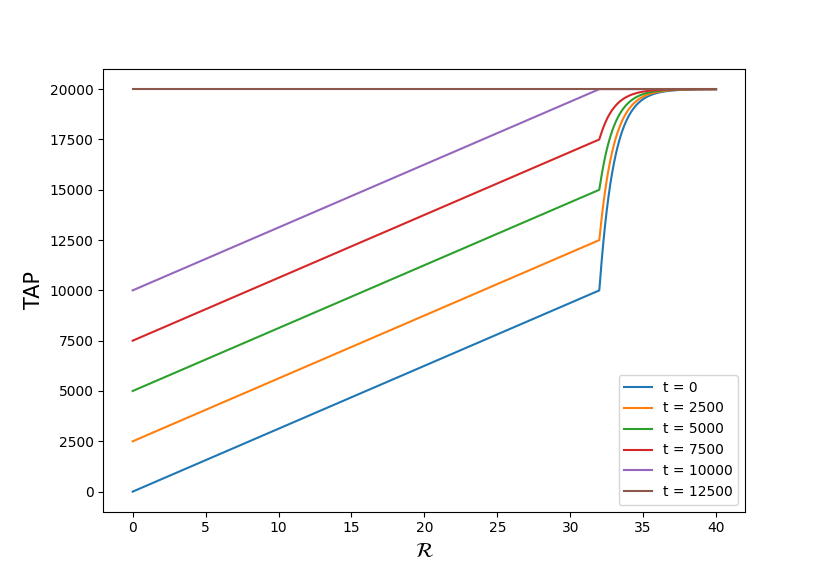}%

	\caption{Variation in TAP with $\mathcal{R}$ for constant t (best viewed in color)}\label{fig:varR}
	
\end{figure}

\begin{figure*}[t]
	\centering

	\begin{subfigure}{0.41\textwidth}
		\centering
		\includegraphics[scale=0.25]{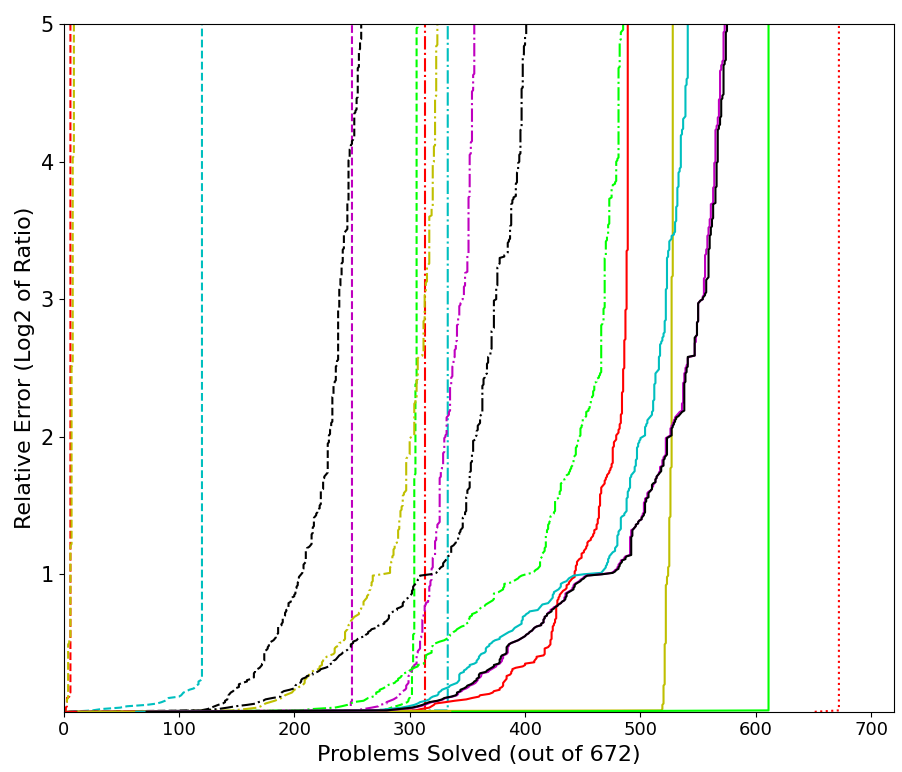}
		\caption{Relative error} \label{fig:error}
	\end{subfigure}
	\hspace*{\fill} %
	\begin{subfigure}{0.075\textwidth}
		\includegraphics[scale=0.25]{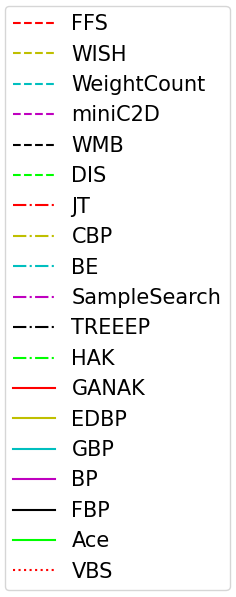}\\\\\\
	\end{subfigure}
	\hspace*{\fill} %
	\begin{subfigure}{0.43\textwidth}
		\centering
		\includegraphics[scale=0.25]{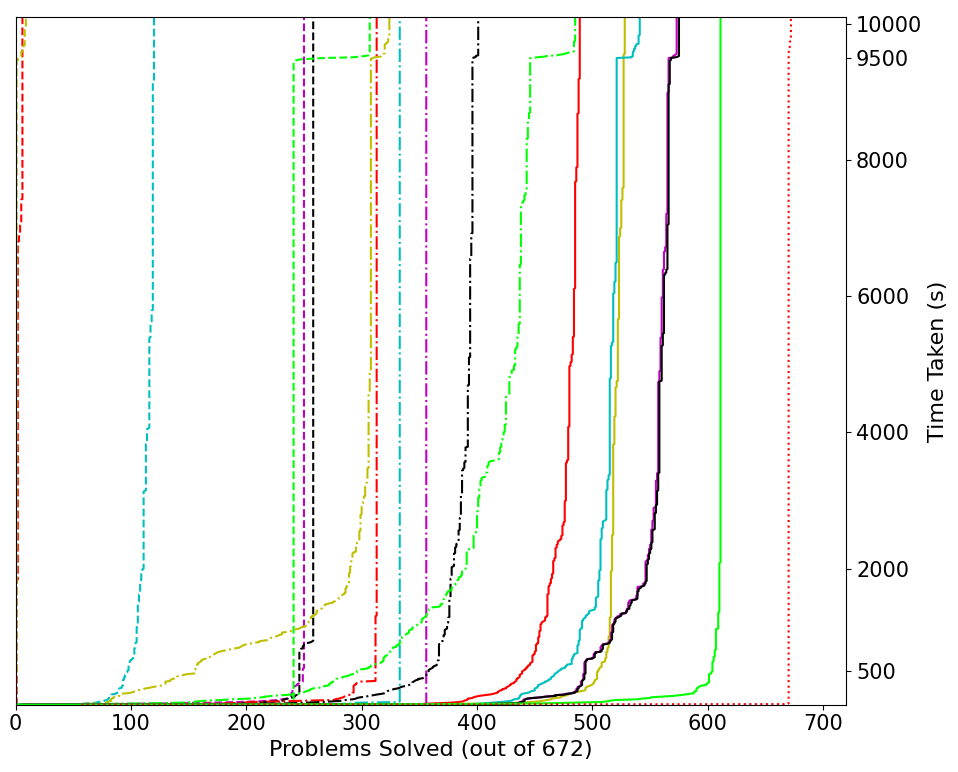}
		\caption{Time taken} \label{fig:time}	
	\end{subfigure}
	
	\caption{Cactus plots (best viewed in color)}
\end{figure*}

\begin{figure*}[t]
	
	\centering
	\includegraphics[scale=0.455]{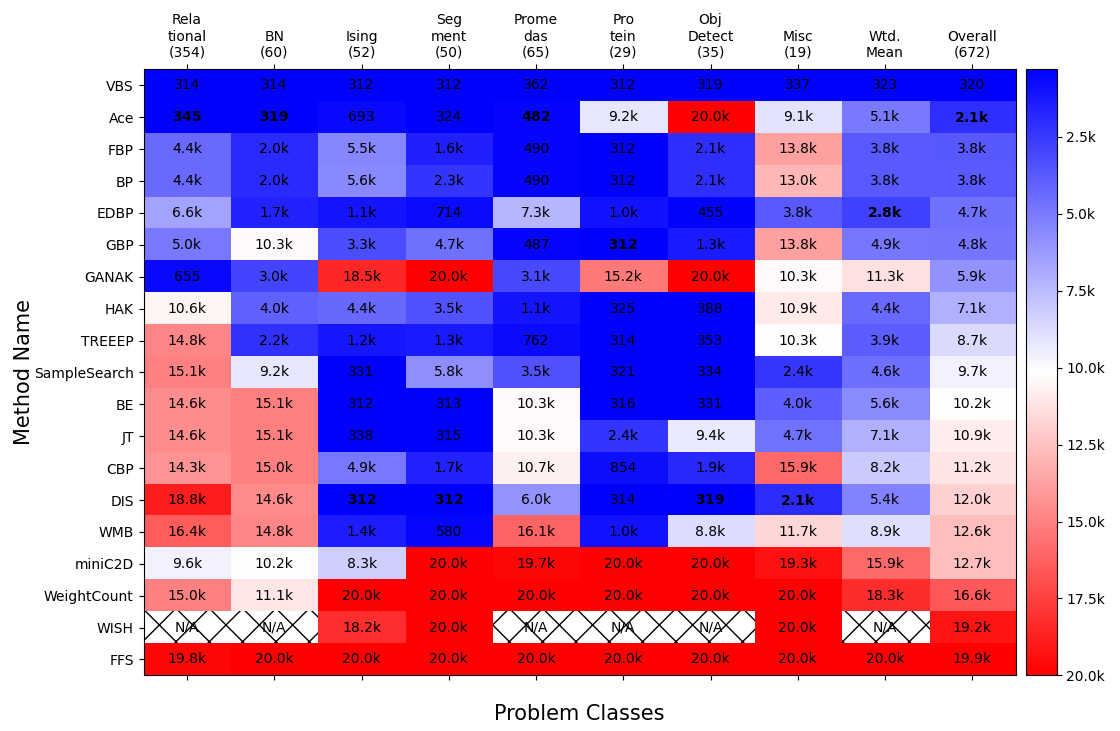}
	\caption{{\tap} score heatmap - method $\times$ problem class}
	\label{fig:penalty_725}	
\end{figure*}

%% file: sections/evaluation.tex
\section{Experimental Evaluation}\label{sec:results}

The experimental evaluation was conducted on a supercomputing cluster provided by NSCC, Singapore\footnote{The detailed data is available at {\protect\url{https://doi.org/10.5281/zenodo.4769117}}}.
Each experiment consisted of running a tool on a particular benchmark on a 2.60GHz core with a memory limit of 4 GB. The objective of our experimental evaluation was to measure empirical performance of techniques along the \textsc{Benchmarks Solved}, \textsc{Runtime Variation}, \textsc{Accuracy} and \textsc{{\tap} score}. 

\subsection{Benchmarks}
Table~\ref{tab:problems} presents a characterization of the graphical models employed in the evaluation on the number of variables ($|\mathbf{X}|$), the number of factors ($|\mathbf{S}|$), the cardinality of variables, and the scope size of factors. 
The evidences (if any) were incorporated into the model itself by adding one factor per evidence. This step is necessary for a fair comparison with methods that use \texttt{.cnf} since they do not take evidence as a separate parameter.

Different implementations and libraries accepted graphical models in different formats: \textit{Merlin}, DIS, SampleSearch, Ace and WISH used the \texttt{.uai} format; \textit{libDAI} used \texttt{.fg} files obtained using a converter in \textit{libDAI}; WeightCount, miniC2D and GANAK used weighted \texttt{.cnf} files obtained using a utility in Ace's implementation; FocusedFlatSAT used unweighted \texttt{.cnf} files.

\subsection{Reliable Algorithms}
To obtain $\hat{Z}$ for 672 benchmarks as described in Section \ref{sec:methodology}, the \textit{reliable} algorithms chosen were Bucket Elimination and miniC2D.

$Z$ computed by Ace and Junction Tree algorithm agree with each other in all the cases when they both return an answer, i.e. the log$_2$ of their outputs are identical upto three decimal places. To verify the robustness of computing $\hat{Z}$ using \textit{reliable} algorithms, we focus on the benchmarks where we can calculate both the true $Z$ and the median of the outputs of \textit{reliable} algorithms. For such benchmarks, the log$_2$ of two values are identical upto three decimal places in $99.3\%$ cases. This implies that by computing $\hat{Z}$ using the approach defined in \ref{sec:benchmarking}, the dataset can be extended effectively and reliably.

\subsection{Implementation Details}

The implementations of all Message Passing Algorithms except Iterative Join Graph Propagation and Edge Deletion Belief Propagation are taken from \textit{libDAI}~\cite{M10}. The library \textit{Merlin} implements Bucket Elimination and Weighted Mini Bucket Elimination~\cite{merlin}. The tolerance was set to $0.001$ in \textit{libDAI} and \textit{Merlin} wherever applicable, which is a measure of how small the updates can be in successive iterations, after which the algorithm is said to have \textit{converged}. For the rest of the methods, implementations provided by respective authors have been used. The available implementation of WISH supports the computation of $Z$ over factor graphs containing unary/binary variables and factors only.

\subsection{Results}
We include the results of a \textit{Virtual Best Solver} (VBS) in our comparisons. A VBS is a hypothetical solver that performs as well as the best performing method for each benchmark.

\subsubsection{\textsc{Benchmarks Solved}}
Table \ref{tab:solved} describes the number of problems solved within a 32-factor accuracy. Abbreviations for the methods are mentioned in parentheses that are also used in the results below. To handle cases when a particular algorithm returns a highly inaccurate estimate of $Z$ before the timeout, we consider a problem \textit{solved} by a particular algorithm only if the value returned is different from the $\hat{Z}$ by a factor of at most 32.

Ace solves the maximum number of problems, followed by Loopy and Fractional Belief Propagations. In problem classes with larger variable cardinality, BP and FBP solve more problems than Ace.
Other methods that exactly compute the partition function solve significantly fewer problems as compared to Ace.
\subsubsection{\textsc{Accuracy}}

Figure~\ref{fig:error} shows the cactus plot comparing relative errors of algorithms. A curve passing through $(x,y)$ implies that the corresponding method could solve $x$ problems with a relative error of less than $2^y$. The curves of the \textit{reliable} algorithms, as defined in Section~\ref{sec:methodology}, are vertical lines as expected, and its X-intercept is a measure of the number of problems solved.

On the other hand, BP and its variants have curves that advance relatively smoothly. This is because these methods cannot provide guarantees for graphical models that do not have a tree-like structure. 
It should be noted that Ace not only returns the exact values of the partition function but also solves the maximum number of benchmarks amongst all algorithms under a 32-factor accuracy restriction.

From the VBS graph, it can be inferred that every problem in the set of benchmarks can be solved with a relative error of less than $2^{0.01}$ by at least one of the methods.

\subsubsection{\textsc{Runtime Variation}}
Figure~\ref{fig:time} shows the cactus plot comparing the time taken by different methods. If a curve passes through $(x,y)$, it implies that the corresponding method could solve $x$ problems with a 32-factor accuracy in not more than $y$ time. The break in the curves at 9500 seconds is due to the soft timeout, and the problems solved after that point have returned an answer within 32-factor accuracy based on incomplete execution.

Vertical curves, such as those of SampleSearch and Bucket Elimination, indicate that these methods either return an answer in a short time or do not return an answer at all for most benchmarks. According to the VBS data, all problems can be solved with a 32-factor accuracy in $<$20 seconds by at least one method. Furthermore,  For 670 out of 672 problems, there exists at least one  method  that can solve the given problem with relative error less than $2^{0.01}$ in time less than 500 seconds.

\subsubsection{\textsc{{\tap} Score}}

We plot a heatmap of the {\tap} score of the algorithms over problem classes in Figure \ref{fig:penalty_725}. `Wtd. Mean' assigns equal weight to each problem class despite its size, whereas `Overall' takes the average of {\tap} score over all the problems.
If method $m$ has the lowest {\tap} score for a problem class $c$, it is indicated in \textbf{bold} on the heatmap. If this minimum {\tap} score is comparable to the corresponding score for VBS, it implies that method $m$ performs better than its counterparts on most problems in class $c$. For instance, in \texttt{ObjDetect} class, DIS has a performance comparable to the Virtual Best Solver.

Despite the lack of formal guarantees, Belief Propagation and its variants have a low overall {\tap} score, and a relatively consistent performance across all classes. Thus, BP is the best candidate to perform inference on an assorted dataset if formal guarantees are unnecessary. Among the exact methods, Ace performs significantly better than others, and it should be preferred for exact inference. 

GANAK, a counting-based method performs well on \texttt{Relational}, \texttt{Promedas}, and \texttt{BN} problems that have a higher factor scope size on an average. However, its {\tap} scores on the classes with a lower factor scope size such as \texttt{Ising}, \texttt{Segment}, \texttt{ObjDetect}, and \texttt{Protein} is high, signaling a poor performance.
The opposite is valid for a subset of methods that do not use model counting, i.e., they perform well on classes with a smaller factor scope size. The methods that show such behavior are JT, BE, and DIS.

\subsection{Limitations and Threats to Validity}
The widespread applications of the partition function estimation problem have prompted the development of a substantial number of algorithms and their modifications that employ various input formats and implementation methods. The sheer diversity makes it impossible to conduct an exhaustive, impartial study of all the available approaches.

To name a few, the K* method~\cite{LSAD05}, the A* algorithm~\cite{LL98}, and a method using randomly perturbed MAP solvers~\cite{HJ12} have not been compared due to the unavailability of a suitable implementation. Also, of note are the recently proposed techniques for weighted model counting that have shown to perform well in the model counting competition~\cite{FHH20}. 
Likewise, benchmarks that could not be converted into compatible formats were not included in the study.

%% file: sections/conclusion.tex
\section{Concluding Remarks and Outlook}\label{sec:conclusion}

Several observations are in order based on the extensive empirical study: First, observe that the VBS has a mean {\tap} score an order of magnitude lower than the best solver. Such an observation in the context of SAT solving led to an exciting series of works on the design of portfolio solvers~\cite{HHS07,XHHL08}. In this context,  it is worth highlighting that for every problem, at least one method was able to solve in less than 20 seconds with a 32-factor accuracy. Also, for every benchmark, there existed a technique that could compute an answer with a relative error less than $2^{0.01}$ in $<$500 seconds. 

Secondly, model counting-based exact techniques are as competitive as techniques without guarantees on the quality of estimates. Coupled with the surprisingly weak performance of approximate techniques, we believe that there is an exciting opportunity for the development of techniques that scale better than exact techniques. 

Finally, the notion of {\tap} score introduced in this paper allows us to compare approximate techniques by considering both the estimate quality and the runtime performance. We have made the empirical data public, and we invite the community to perform a detailed analysis of the behavior of techniques in the context of structural parameters such as treewidth, the community structure of incidence and primal graphs, and the like. 

%% file: sections/acks.tex
\section*{Acknowledgments}
The authors would like to sincerely thank Adnan Darwiche, and the anonymous reviewers of IJCAI-20 and AAAI-21 for providing detailed, constructive criticism that has significantly improved the quality of the paper. 

This work was supported in part by National Research Foundation Singapore under its NRF Fellowship Programme [NRF-NRFFAI1-2019-0004] and AI Singapore Programme [AISG-RP-2018-005],  and NUS ODPRT Grant [R-252-000-685-13]. The computational work for this article was performed on resources of the National Supercomputing Centre, Singapore (\href{https://www.nscc.sg}{https://www.nscc.sg}).